\documentclass[conference]{IEEEtran}
\IEEEoverridecommandlockouts
\usepackage{cite}
\usepackage{amsmath,amssymb,amsfonts}
\usepackage{algorithmic}
\usepackage{graphicx}
\usepackage{textcomp}
\usepackage{xcolor}
\def\BibTeX{{\rm B\kern-.05em{\sc i\kern-.025em b}\kern-.08em
T\kern-.1667em\lower.7ex\hbox{E}\kern-.125emX}}
\usepackage[numbers,sort&compress]{natbib}
\begin{document}

\title{Gen-AI for User Safety: A Survey\\
\thanks{* These authors contributed equally.}
}

\makeatletter
\newcommand{\linebreakand}{%
  \end{@IEEEauthorhalign}
  \hfill\mbox{}\par
  \mbox{}\hfill\begin{@IEEEauthorhalign}
}
\makeatother

\author{\IEEEauthorblockN{Akshar Prabhu Desai \footnotemark{*}}
\IEEEauthorblockA{akshard@google.com}
\and
\IEEEauthorblockN{Tejasvi Ravi \footnotemark{*}}
\IEEEauthorblockA{ravitejasvi@google.com}
\and
\IEEEauthorblockN{Mohammad Luqman}
\IEEEauthorblockA{moluqman@google.com}
\linebreakand
\IEEEauthorblockN{Mohit Sharma}
\IEEEauthorblockA{mohitzsh@google.com}
\and
\IEEEauthorblockN{Nithya Kota}
\IEEEauthorblockA{nithyakota@google.com}
\and
\IEEEauthorblockN{Pranjul Yadav }
\IEEEauthorblockA{pranjulyadav@google.com}
\linebreakand
\IEEEauthorblockA{Google}
}

\maketitle

\begin{abstract}

Machine Learning and data mining techniques (i.e. supervised and unsupervised techniques) are used across domains to detect user safety violations. Examples include classifiers used to detect whether an email is spam or a web-page is requesting bank login information. However, existing ML/DM classifiers are limited in their ability to understand natural languages w.r.t the context and nuances. The aforementioned challenges are overcome with the arrival of Gen-AI techniques, along with their inherent ability w.r.t translation between languages, fine-tuning between various tasks and domains.

In this manuscript, we provide a comprehensive overview of the various work done while using Gen-AI techniques w.r.t user safety. In particular, we first provide the various domains (e.g. phishing, malware, content moderation, counterfeit, physical safety) across which Gen-AI techniques have been applied. Next, we provide how Gen-AI techniques can be used in conjunction with various data modalities i.e. text, images, videos, audio, executable binaries to detect violations of user-safety. Further, also provide an overview of how Gen-AI techniques can be used in an adversarial setting. We believe that this work represents the first summarization of Gen-AI techniques for user-safety. 

\end{abstract}


\section{Introduction}

%%% ADD SOME CITATION %%%% 

Machine Learning and data mining techniques are used across domains to detect violation of user safety. Examples include classifiers used to detect whether an email is spam or a web-page is requesting bank login information. However, existing ML/DM classifiers are limited in their ability to understand natural languages w.r.t the context and nuances. The aforementioned challenges are overcome with the arrival of Gen-AI techniques, along with there inherent ability w.r.t translation between languages, fine-tuning between various tasks and domains.

In this manuscript, we provide an overview of the various  domains across which user safety can be violated. In particular, we provide more information on how Generative Artificial Intelligence (Gen-AI) techniques can be used towards reduction of egregious abuse (e.g. phishing, malware, anomaly detection, counterfeit, fraud prevention), misinformation and disinformation (e.g. fake news , deepfake detection), increase in content moderation, awareness about mental health (e.g. cyber-bullying  prevention, crisis support) and towards robust physical safety (e.g. accessibility, autonomous systems).

Further, we discuss how Gen-AI techniques can be used across various data modalities. In particular, we present how Gen-AI techniques can be used to detect user-safety violations in text and rather outperform all previous techniques w.r.t NLP tasks such as entity recognition, question answering and sentiment analysis. Further, Gen-AI techniques with their inherent ability to parse and understand images provides an easy mechanism to detect image manipulation, deepfake detection. The advantages of Gen-AI techniques go beyond text and images to other data modalities such as videos, audio and executable binaries. 

We also discuss how Gen-AI techniques can be used in an adversarial setting. In particular, we present how these techniques can be used to attack at scale (e.g. mass spam). Further, the attacks become more intelligent as these techniques can target humans more effectively and engage with humans with a similar cognitive capacity. Gen-AI techniques along with reinforcement learning techniques can be used to create more sophisticated attacks while using feedback from the last failure. Further, Gen-AI techniques can also make these attacks look very personalized (e.g. deep-fakes) and second order effects (e.g. Gen-AI imitating human sounding text).

The organization of the paper is as follows. In section \ref{usd}, we provide a comprehensive overview of the various user safety domains, where Gen-AI techniques can be applied. In section-3, we discuss the various data modalities across which Gen-AI techniques can be utilized to protect user safety violation. In Section-4, we present, how Gen-AI techniques can also be used in an adversarial setting. In Section-5, we present our opinion on what does the future look like for Gen-AI techniques. Finally, in section-6, we conclude this manuscript. 

%----------------------------------------------------------------------------------------------------------------------------%
\section{User Safety Domains}\label{usd}
Gen-AI techniques can significantly enhance user safety in both digital and physical environments. It can proactively address risks, offer timely assistance, and empower individuals with personalized tools. Within digital realm, it can detect fraud, flag instances of harmful content, cyberbullying, and predatory behavior. In the physical realm, Gen-AI techniques can help improve accessibility via wearables, contribute to mental and physical health by providing timely access to online resources and perform crisis intervention if necessary. Below we explore and detail the research that enhance user safety in these two realms.

\subsection{Digital Realm}

\subsubsection{Online Threat Protection} 

Phishing is a type of socially engineered cyber-attack where bad actors try to trick users into giving up their personal information by pretending to be a trustworthy source either via email and phishing websites \cite{targeted-threat-index}. The work by Koide et al. \cite{Koide2023DetectingPS} showed that Gen-AI techniques lowers the barrier to deploy systems that detect, mitigate and deter new phishing attacks by utilizing their broad knowledge base, multi-modal and multi-lingual capabilities, which otherwise would require multiple different classifiers \cite{marchal2016knowphishnoveltechniques}. Further, Koide et al. \cite{koide2024chatspamdetectorleveraginglargelanguage} presented Gen-AI techniques that can provide user with detailed reasons for why certain email or website is highlighted for phishing. Ai et al. \cite{ai2024defendingsocialengineeringattacks} proposed advanced strategies like Retrieval-Augmented-Generation (RAG), which can detect sophisticated phishing attacks that involve longer multi-turn interaction between bad actor and unsuspecting users.

Malware is malicious software, that includes viruses, worms, and Trojan horses, deliberately designed to compromise computer systems, servers, or networks. These attacks usually lead to data breaches (e.g., via spyware), system damage, or unauthorized control of devices (e.g., through adware and ransomware).  Gen-AI techniques help enhance existing strategies for malware identification and alert generation that have traditionally employed machine learning and deep learning models \cite{Bensaoud_2024}\cite{nagashree2018early}. Ferrag et al.\cite{ferrag2024generative} showed that with advanced techniques, such as Half-Quadratic Quantization (HQQ), Direct Preference Optimization (DPO), GPT-Generated Unified Format (GGUF), Quantized Low-Rank Adapters (QLoRA), and Retrieval-Augmented Generation (RAG), Gen-AI can be leveraged for more effective threat detection and response. In AppPoet, Zhao et al. \cite{zhao2024apppoetlargelanguagemodel} demonstrated a system that uses multi-view prompt engineering to detect and produce a detailed diagnostic report for android malware. Similarly, Wang et al. \cite{10.1145/3663408.3663424} in their work ShieldGPT, showed that via prompt engineering, Gen-AI has the potential to defend against Distributed Denial of Service (DDoS) attacks and provide comprehensible explanation and detailed mitigation instructions specific to an attack.

\subsubsection{Misinformation Detection}
Fake News is false or misleading information presented as news. The proliferation of misinformation across social media platforms and news outlets presents a significant threat in the digital age. The sheer volume of online content renders manual fact-checking impractical and this is where Gen-AI can help. 

Zhang et al. \cite{Zhang2023TowardsLF} proposed a hierarchical prompting method that outperforms state of the art fully-supervised approach for news claim verification. Further, the work by Yue et al. \cite{Yue2024EvidenceDrivenRA} proposed a retrieval augmented response generation system to combat online misinformation and generated counter-misinformation responses based on the scientific evidences. Furthermore, work by Xuan et al. \cite{xuan2024lemmalvlmenhancedmultimodalmisinformation} showed how Gen-AI can utilize external knowledge bases for information verification.

Gen-AI are proving effective in detecting fake news, both independently and in conjunction with specialized Small Language Models (SLMs) \cite{Hu_2024_fake_news}. Notably, Gen-AI can identify misinformation generated not only by humans but also by automated systems \cite{su2024adaptingfakenewsdetection}. Their capabilities extend to multimodal misinformation encompassing text, images, and videos \cite{Wang2024MMIDRTL}.

Deepfakes are synthetic media where an entity in an existing image or video is replaced with another entity's (usually a person) likeness using artificial intelligence techniques. Recent advances in Gen-AI has enabled creation of extremely high fidelity personalized content like images, audio and video. At the same time, several techniques has been developed to detect such deepfakes across different modalities.
Chen et al. \cite{pmlr-v235-chen24ay} proposed a framework called Diffusion Reconstruction Contrastive Learning (DRCT), that generated hard samples by high-quality diffusion reconstruction and adopted contrastive training to guide the learning of diffusion artifacts. They showed that detectors enhanced with DRCT achieve over a 10\% accuracy improvement in detecting diffusion generated images. Further, Zhang et al. \cite{Zhang2021DetectingDV} proposed a Temporal Dropout 3-dimensional Convolutional Neural Network (TD-3DCNN) to detect deepfake videos model on public deepfake datasets that surpassed state of the art performance on detecting deepfake videos.

\subsubsection{Content Moderation}
Content moderation is critical to maintaining the integrity of online platforms and to keep the members safe from harmful content, misinformation, and disinformation, and to comply with legal and policy standards. Content moderation is a growing challenge as the platforms scale and sheer volume of content to be moderated can't be handled by humans alone. Diversity of content in terms of language and modality and evolving nature of harmful content itself add an additional challenge.
The in-context learning \cite{dong2024surveyincontextlearning} ability of Gen-AI techniques have been used to encode online platform policy in a prompt to detect whether content violates it or not \cite{10.1145/3613905.3650828}. The multi-modal reasoning capability of Gen-AI now allows us to capture the necessary context spread across text, images and video (short and long form) components to detect policy violation.
Ahmed et al. \cite{ahmed2023potentialvisionlanguagemodelscontent} proposed a model that outperforms previous work conducted on the Malicious or Benign (MOB) benchmark for video content moderation. They also highlight the importance of providing context to content moderation prompts to improve the performance.

\subsection{Physical Realm}
This section focuses on user safety enhancements that can impact a user outside of the digital environment like accessibility/ mobility improvements, crisis support and improvements to mental health and well being.

During emergencies, support systems are overwhelmed. Gen-AI powered chatbots can aid in \textbf{crisis support} by providing rapid targeted information, enhancing communication, offering emotional support, and improving preparedness. Advanced frameworks that leverage Gen-AI can be utilized to help the support systems by understanding user needs and creating workflows for government agencies as documented in the survey by Rieskamp et al. \cite{rieskamp2023genai}. Further, the work by Otal et al. \cite{10607148,10605553}, explored using fine tuned models like LLama2 to assist users with simple instructions while informing authorities with summarized and accurate information. 

Gen-AI techniques can enhance accessibility, enabling safer navigation in the world for visually impaired people. In VisionGPT, Wang et al. \cite{wang2024visiongpt} showcased a system that takes in real time video via camera captured frames and provides a concise audio description that enabled safer navigation. 

Counterfeit goods are fake products designed to look genuine, like branded items. We include them in the physical realm as they involve a non digital entity that the user interacts with. Production and distribution of counterfeit goods infringe on intellectual property rights which can cause serious damage to consumer health and safety (Counterfeit medicines, cosmetics, or safety equipment) and brand reputation, eventually resulting in losses for legitimate businesses. According to the National Crime Prevention Council around \$2 trillion worth of counterfeit products are sold to consumers annually \cite{cbsnews_2024_counterfeit}. At the time of writing, there isn't much research around using Gen-AI techniques for counterfeit detection, however, there are relevant work in identifying counterfeits with Generative Adversarial Networks (GANs). The generative component mirrors the counterfeiter's role, while the discriminator functions as the detective, identifying and rejecting fraudulent outputs. Some examples include, a combination of external attention GAN with deep convolutional neural networks (CNNs) developed by Peng et al. \cite{peng2023combining} to identify counterfeit luxury handbags and GANs for credit card fraud detection as shown by Wang et al \cite{wang2024novel}.

\subsubsection{Mental Health and Well-being}
Mental health and well-being is a topic that is central to user safety and Gen-AI can be applied to detect, intervene, prevent and provide timely support to ensure users mental well being. VITA \cite{spitale2023vita}, a multi-modal Gen-AI based system for mental well being, allows robotic coaches to autonomously adapt to the coachee's behaviours from features like facial valence and speech duration. Using these signals, it delivers adaptable coaching exercises to promote mental well being. Along with useful intervention activity recommendations, Gen-AI based agents that are anthromorphic are able to foster relational warmth and can prove to be more effective\cite{wu2024i}.

Gen-AI can be trained to detect and flag \textbf{cyberbullying} instances in online interactions, which often results in significant psychological distress for the victims, allowing for faster response and support for victims. Vanpech et al.\cite{10499678} proposed a system to identify cyberbullying via images by feeding the image as input to GPT-4 to generate description metadata, which was then provided to a custom trained Gen-AI model that classified the images to detect if it's used for bullying. Gen-AI can also be used to augment training data to improve existing classifiers. For example, in the work by Jahan et al. \cite{jahan2024comprehensive} GPT-3 based data augmentation showed 0.8\% improvement in classification F1 score for hate speech detection tasks.

\textbf{Predatory behavior detection} is a critical research area for social media platforms to ensure user safety, especially for vulnerable populations. Gen-AI can analyze patterns in text and images to identify grooming behaviors or attempts to solicit explicit content from minors. This can help platforms intervene proactively and protect users from online predators. Nguyen et al. \cite{nguyen2023finetuning} discusses how fine tuned models are able to detect online predators better than out of the box foundational models. In particular, they used a LLama 2 model which was fine-tuned using LORA.

%----------------------------------------------------------------------------------------------------------------------------%
\section{Data Modalities} \label{data-modalities}

This section explores how Gen-AI impacts user safety across different data modalities.

\subsection{Text}
Large language models, such as GPT-4 , Gemini and LLaMA have demonstrated outstanding performance across downstream NLP (Natural Language Processing) tasks (e.g. text classification, named entity recognition, translation, question answering and sentiment analysis) \cite{zhang2024pushing}. The inbuilt context from pre-training enables transformer based models like RoBERTa \cite{liu2019roberta} to perform well even on tasks that were difficult earlier, such as sarcasm detection \cite{potamias2020transformer}.  In particular for user safety, pre-trained transformer models achieve remarkable performance in hate-speech detection, detecting spam, fake news and fake reviews. Further, Keyan et al. demonstrated that well crafted reasoning prompt can effectively capture the context of hate speech by fully utilizing the knowledge base in Gen-AI models, significantly outperforming existing techniques \cite{guo2023investigation}. Using retrieval augmented generation (RAG), these models are able to perform fact verification \cite{lewis2020retrieval}, and are also able to detect fake news written in high quality journalistic style \cite{wu2024fake}.

Gen-AI techniques are also able to overcome content moderation to low resource languages as well as to multilingual text. In particular, multilingual-BERT and XLM-RoBERTa, each of which have been pre-trained on 100+ languages have been used to perform well on hate speech and hostility detection with multilingual input \cite{yadav2024code, sai2021stacked}. Further, they are also being used to generate annotated data for training models for low resource languages. Additionally,  annotated data generated by Gen-AI techniques are found to be on par with human annotators and can be done at a fraction of time and cost \cite{kholodna2024llms}. 

\subsection{Images}
Following the breakthroughs in large language models, large multimodal models (LMMs) such as GPT-4v \cite{yang2023dawn} extend these capabilities to other data modalities, with images being of particular focus. User safety in the domain of images involves detecting harmful images, images that violate platform policies, flagging sensitive media and/or assisting users in detecting machine generate image from real images. 

\subsubsection{DeepFake detection} 
Easy availability of image generation tools such as Midjourney, StableDiffusion \cite{rombach2022high}, Dall-E \cite{betker2023improving} and others have led to proliferation of fabricated images online. These models generate hyper realistic images that are not easily distinguished from real images. 

Existing techniques for detecting deepfake images are mostly formulated as binary classification problems and fall into three major categories - identifying inconsistencies exhibited in the physical/physiological aspects in the DeepFake images, methods using signal-level artifacts introduced during the synthesis process, or directly training a classifier on real and DeepFake samples. Another shortcoming is that classifiers trained to detect images generated from one class of generative models (e.g., GAN) fail to generalize on other generative models (e.g., diffusion models)  \cite{ojha2023towards}. The same techniques that have resulted in the success of image generation have also been employed to detect deepfake images. Most recently, using a pre-trained CLIP-ViT model to learn image features followed by a classifier to detect fake images sets new state of the art and also generalizes across different types of generative models \cite{ojha2023towards}.

\subsubsection {Harmful images detection}
Recent advancements in vision-language models have improved the ability to detect harmful images, making them useful for content moderation \cite{guo2024moderating} \cite{vishwamitra2021towards}. Models like VinVL \cite{zhang2021vinvl}, which leverage transformers and attention mechanisms, can capture complex relationships between visual elements and generate accurate, contextually relevant captions. Further, large-scale multi-modal pre-training on massive datasets of images and text, such as CLIP (Contrastive Language-Image Pre-training) \cite{li2021supervision}, has enhanced the models' ability to connect visual and textual information \cite{radford2021learning}. This is crucial for identifying harmful images, especially those containing hate speech or offensive symbols, which often rely on the interplay between visual and textual elements.

PaLI-X \cite{hu2024visual}, a vision language model trained using latest techniques of instruction-tuning and distillation, outperforms all prior Vision Language Models (VLMs) achieving state-of-the-art performance across complex vision tasks including the hateful memes challenge, which seeks to identify multi-modal hate speech. 

A comparative study of different image safety classifiers performed \cite{qu2024unsafebench} observed that VLMs can identify a wider range of unsafe content, with GPT-4v \cite{openai2024gpt4technicalreport} being the top performing model for this use case. 

\subsubsection{Safety applications}

VLM's nuanced understanding of a scene has applications in defect recognition and safety equipment recognition \cite{yang2023dawn}. These models are being used to detect safe pedestrian crossing \cite{hwang2024safe}, adherence to workplace safety guidelines \cite{chen2024vision}, and evaluation of construction site safety among others \cite{tsai4819831construction}. 

\subsection{Videos}

Consumption of video content has been on steady increase especially due to social media and streaming platforms \cite{cheng2007understanding}. User safety issues in videos could be the presence of harmful, age inappropriate, violence , copyright violations \cite{zhang2018end} or deepfakes. 

Technology companies have built sophisticated solutions to understand video content and protect users from these threats \cite{kandakatla2016identifying}. Following subsections outline specific challenges related to user safety for different types of video content.

\subsubsection{Human generated videos}
Video forensics field has been using Gen-AI tools to extract information from videos to detect threats. This includes processing individual frames as images and detecting objects and actions in those frames, converting audio to text and processing that text with Gen-AI techniques for further analysis. Bi-Long Short Term Memory (Bi-LSTM) machine learning model combined with Convolutional Neural Networks (CNNs) are often used to detecting violence and other harmful behavior in video content. Sentiment analysis of the audio in the video too can be used to detect harmful videos.  

Gen-AI breakthroughs now allow people to generate videos simply by using text prompts. Models like Sora have achieved remarkable results. However, this unrestricted ability to create videos introduces user safety concerns. 

These generated videos present two main challenges.
\begin{itemize}
    \item Detecting generated videos that claim to be real (deepfakes)
    \item Detecting harm in generated videos
 \end{itemize}
 
Pang et al. \cite{pang2024towards} showcased defense mechanisms to prevent the generation of unsafe videos via their approach called Latent Variable Defense (LVD), which works within the model's internal sampling process. LVD achieved 0.90 defense accuracy while reducing time and computing resources by 10x when sampling a large number of unsafe prompts. Further, Rana et al. \cite{rana2022deepfake} provided an comprehensive overview of existing research works for deepfake detection. 

Detecting harm in generated videos is another important research problem. Since generated videos are not bound by the physics of real world, generated videos can cleverly twist certain elements of the video to evade detection of standard algorithms\cite{kingra2023emergence}.

\subsubsection{Live streaming}
Live video feeds is a popular form of video content. Besides social media live streams, security camera footage, traffic camera feeds, live sports and gaming etc. are important sources of live stream videos \cite{liz2024generation}. 

Fan et al. \cite{fanreal} discuss a real time deepfake identification framework to handle abuse in live streams. Further Liz-Lopez et al. \cite{liz2024generation} provides an extensive survey of technologies that can be used for detecting harmful multi-modal content in live streams. Furthermore, Gupta et al. \cite{gupta2023real} proposed quantum machine learning as another potential approach for processing live streams. 

\subsection{Audio}

Traditional machine learning techniques are limited in their ability to handle fabricated and harmful audio content. This inability usually stems as traditional methods rely on the extraction of acoustic features in the spectral domain. However, with the advent of latest tools for generating natural sounding voice, traditional methods are bound to fail.

Gen-AI techniques are able to overcome this by helping in dataset generation for advancing research. Datasets such as FakeAVCeleb \cite{alali2024review, khalid2021fakeavceleb} and Joint Audio-Visual Deepfake \cite{zhou2021joint} improve deepfake detection research by including video deepfakes with corresponding synthesized, lipsynced audio tracks or by integrating audio alteration. Further efforts, such as WaveFake \cite{frank2021wavefake}, which contains 100K+ generated audio samples contribute substantially to building resources capable of addressing the problem of detecting deepfakes. PolyGlotFake \cite{hou2024polyglotfake}, a multi-modal and multi-lingual deepfake dataset covers 7 languages, and  MLAAD (Multi-Language Audio Anti-Spoofing Dataset)  expands audio spoofing dataset to 23 languages \cite{muller2024mlaad}. 

% Easy availability of tools like Wavenet, MelNet, Char2Wave and WaveGlow have made generating human sounding speech is an easy task \cite{van2016wavenet} \cite{ vasquez2019melnet} \cite{ prenger2019waveglow}. Tools such as Wav2Lip \cite{prajwal2020lip} and Obamanet \cite{kumar2017obamanet} make it easy to generate very realistic lip synced videos. 

Gen-AI techniques are also able to detect hate speech in verbal data.  In particular, conformer model, an architecture combining convolutional networks and transformers, improves automatic speech recognition with a word error rate (WER) of less than 2\% \cite{gulati2020conformer} . Further, generative models are helping to fill the lack of audio only datasets. For example, An et al. used text to speech models to generate an audio only dataset, and a BERT based model for explainable hate speech detection directly from audio files \cite{an2024investigation}.

\subsection{Code}

Gen-AI do not simply generate code or superficially understand its context. They clearly demonstrate the ability to process, identify the relevant parts, and operate on them. This capability is demonstrated in detecting malicious code, that is deliberately obfuscated to prevent this from happening. This makes them effective in de-obfuscating malicious scripts \cite{deobfuscate-malicious-code}, cyber threat detection \cite {cyber-threat-detection}, and even understanding minified code \cite{unminify-code}. 

Chuanbo et al. \cite{hu2024multimodal} systematically leverage ChatGPT-4 to process multimodal app data (i.e.,
textual descriptions and screenshots) to determine maturity rating of an app to keep children safe from age inappropriate apps.

\section{Adversarial Gen-AI}

In this section we analyze how Gen-AI technologies are impacting the adversaries of user safety.  Earlier sections detail the well-understood threats to user safety. However, to better detect and deter these threats, we must better understand how hostile actors might use Gen-AI techniques in adversarial settings.

\subsection{Safety Violations at Scale}

Online fraud poses a major threat to user safety, with bad actors frequently aiming for large-scale disruption. Technology facilitates these attacks by enabling the mass distribution of emails and text messages. However, a limiting factor for these bad actors lies in their access to limited resources to conduct intelligent conversations with each victim. Gen-AI techniques allows bad actors to overcome this crucial limitation \cite{basit2021comprehensive}. 

By using Gen-AI technology, they can reduce human involvement in large-scale operations thereby decreasing the time and cost of their attacks. In particular, bad actors can craft more convincing communications and tailor them to each user, making detection by spam and threat detection algorithms more difficult \cite{de2023unethical}. Additionally, Gen-AI enables large-scale content generation\cite{wu2024fake}, facilitating the creation of fake news websites and blogs that produce convincing, yet deceptive, content \cite{de2023unethical}. Further, foreign language content creation (i.e. geo-targeting) which was a barrier for bad actors can be easily overcome using Gen-AI technology \cite{mitra2024world}.

% Thus, Gen-AI has an impact on bottlenecks described in \ref{table:attacks}. Gen-AI might allow bad actors to operate at higher scale with higher quality of deceptive content and might also create new distribution channels of communication such as chat-bots, content websites and generated videos. 

\subsection{Safety Violations with Feedback}
Considering the adversarial nature of abuse, bad actors can use sophisticated Gen-AI techniques to make more effective attacks using explicit feedback from past attempts.  

In particular, reinforcement learning can help bad actors create better user-violation models that help them create targeted and effective campaigns \cite{beckerich2023ratgpt} to violate the safety of a user. Additionally, bad actors have access to data from previous attacks, which is often not accessible to the security researchers. Mujumdar et al. \cite{majumdar2024beyond} show that this data is often utilized by bad actors to improve the effectiveness of their attacks.

Captcha is a common technique used by social media and other applications to prevent abuse from bad actors using automated systems. However, modern Gen-AI techniques have enabled bad actors to develop sophisticated solvers thereby overcoming the barriers posed by captcha. Ye et al. \cite{ye2020using} discussed how the availability of human labeled data enabled researchers to create sophisticated solvers.

\subsection{Safety Violations with Personalized Content}

Gen-AI techniques have improved bad actor’s ability to generate highly personalized content. For example, Gen-AI techniques with a large context window can technically look at a victim’s available information and create personalized\cite{guo2022generating} email campaigns, websites, calls or even video messages for phishing or financial fraud based scams. Further, Gen-AI allows bad actors to turn a normal phishing attack into a spear phishing \cite{phishing} attack where instead of targeting a large group of people with similar messages, a highly targeted strategy is crafted to target specific individual or individuals. 

Gen-AI techniques can also be used for pre-texting \cite{ansari2020scammer} based scams. In this form of attack a bad actor designs an attack plan for their target, using a story around the facts they know about the individual. The end goal of this story is eventually to scam the victim using prior personalized information. This is a complex form of personalized scam and traditionally requires lot of resources from bad actor. However, with Gen-AI technique, such level of personalized abuse is easier to achieve.

%This involves complex websites that look legitimate but really aren't. Gen-AI generated apps that seem high quality but are really fake, deepfake audio, video and pictures etc. can be used to generate highly personalized content to target either specific individuals or narrow groups. In modern conflicts we have already seen both sides using AI to generate realistic looking content to target specific audiences. 

% Might create scare ?? Lets avoid it. Deepfakes are the fake or manipulated videos or pictures of an individual. Once deepfake generation is cheaper and better, ransom campaigns can be run to blackmail people with their deepfake generated videos. Deepfakes are fabricated or altered videos or images of an individual. As deepfake technology becomes more accessible and sophisticated, it could facilitate blackmail campaigns, where individuals are extorted using deepfake videos generated to appear authentic.

\subsection{Second order attacks} \label{second-order-attacks}
Second order attacks are scenarios where the human behavior is manipulated to achieve a certain end result by fake news campaigns and/or social media activity. Synthetic reality \cite{sythenticreality} is an example of a second order effect which includes misinformation or targeted campaigns by bad actors. In such attacks it is not clear who is the victim; but as long as some people fall prey to such activity the bad actors benefit. Attacks like these are extremely well coordinated and require complex infrastructure. It involves creating fake news websites, social media bots and even human users to amplify a certain type of message to make it sound very real. 

Another form of second order effect attacks is where bad actors develop a long term plan to trick the world’s latest Gen-AI technologies. Bad actors develop a long term plan to create synthetic reality on the internet through various specially setup websites, synthetic social media posts and similar mechanisms. Through this approach they poison the data on which Gen-AI techniques are trained and built \cite{pathmanathan2024poisoningrealthreatllm}. 

\subsection{Gen-AI Violations}

AI safety is an evolving field where in models have inbuilt mechanism to ensure that they are not being used for user-harm \cite{zhou2024easyjailbreak, chua2024ai}. However, when such inbuilt safety mechanisms in the model are broken, it is referred to as jail-breaking the model\cite{jailbreak1}. Jail-breaking is an active field of research as well as an area where regulations are being actively sought \cite{ romero2024generative}. Li et al. \cite{li2024lockpicking} proposed a system called JailMine, which is able to overcome the defensive measures of Gen-AI techniques, even when Gen-AI models are frequently updated and incorporated with advanced defensive measures.

Majumdar et al. \cite{majumdar2024beyond} present a detailed survey of how malicious actors might be gaining strategic advantage with the help of Gen-AI techniques despite the AI safety barriers on these models. In particular, they discussed how Gen-AI techniques can be used towards the manipulation of public opinion through the generation of deceptive content, the orchestration of social engineering attacks using sophisticated language-based techniques, and the ease of cybercrimes such as phishing, password cracking and malware propagation. 

\section{Section: Future Prospects of Gen-AI}

With the recent advancements in Gen-AI techniques, we expect to see exciting developments that improve user safety, and this section describes some of these areas.

\subsection{Content understanding}

Content understanding is key to ensuring user safety. Traditionally, human moderators, supported by machine learning technologies, have analyzed content to ensure user safety. This analysis often uses some form of Reinforcement Learning through Human Feedback (RLHF). This has been a challenge due to limited availability of training data \cite{franco2023analyzing}. 

Gen-AI techniques offer complex models that identify content violating user-safety. Institutions leveraging Gen-AI for user safety, no longer need to develop their own content understanding models. Further, Mohammed et al. \cite{mahomed2024auditing} demonstrated that it can be done by off-the-shelf models with some fine tuning or prompt engineering. Gen-AI techniques can also be utilised for minority classes and outliers as they are much better represented in the training data of Gen-AI as compared to the internal data of an institution \cite{franco2023analyzing, htet2024chatgpt}.  Chaudhary et al. \cite{chawdhury2024content} provides an example of how a carefully crafted prompts and an off-the-shelf Gen-AI API's alone can produce excellent content moderation service. The availability of general-purpose models allows for the deployment of sophisticated Gen-AI technologies at a lower cost and without specialized content understanding. \cite{qiao2024scaling}

\subsection{Foundation Ensembles}
Gen-AI techniques enables the availability of sufficiently large and complex models that can handle many business use cases thereby replacing multiple smaller specialized models. For example, a single model can now detect phishing \cite{gallagher2024phishing}, spam, fraud and analyze sentiment within a text. This simplification streamlines the training, fine-tuning, and deployment of these models thereby accelerating development and deployment processes. This reduction in technical stack then results into deprecation of system complexity and decrease in associated costs.   

Detoxbench \cite{chakraborty2024detoxbench} is a benchmark that measures the ability of LLMs to detect various forms  of user safety violations such as toxic content, phishing, spam and fraudulent job postings. 

\subsection{Harnessing multi-modality for better user safety}

Section \ref{data-modalities} discussed the user safety challenges across data modalities and how Gen-AI impacts them. Future user-safety research using Gen-AI should focus on harnessing the multi-modality of data. This research should analyze text, images, audio, and video data holistically for better user safety, rather than treating them as independent inputs.\cite{park2021correspondence}

Yang et al. \cite{yang2023auto} showed that auto-insurance fraud detection is improved with the help of multi-modal models rather than analyzing only structural data. Further, Goyal et al.  \cite{goyal2023detection} hypothesized that there would be more applications of multi-modal AI in improving user safety. Furthermore, Akhare et al. \cite{akhare2024machine} provided a survey of machine learning techniques in area of user safety and discuss that multi-modal multi-objective evolutionary algorithms are more scalable and more precise than other traditional methods to analyze content to enhance user safety.

\subsection{Gen-AI and second order harm prevention}

Future user safety research will likely focus on preventing the "second-order effects" described in section \ref{second-order-attacks}. Research like \cite{khaddage2019good} have shown that long term psychological and financial harm is possible due to widespread Gen-AI usage. For example, Dewite \cite{dewitte2024better} shows that extensive usage of chatbots might be harmful in long term even though independently specific conversations with the chatbot might appear harmless. Markus et al. \cite{anderljung2023protecting} shows that how certain AI uses might be perfectly normal independently but when chained together can create a misuse chain and how this might be detected. 

User safety researchers might have to take more holistic approach towards user safety instead of restricting themselves with specific methods or modalities \cite{chen2024trustworthy}. Kranz et al. 
\cite{kranz2024generative} demonstrated how  AI-copilot can be provided to humans in their interactions with robots to ensure safety, by flagging unusual behavior. 

We anticipate the widespread adoption of co-pilots that monitor users' online behavior across platforms. These co-pilots will provide a comprehensive safety mechanism, protecting users from a wide variety of second-order effects stemming from the pervasive use of Gen-AI in everyday life. Such technologies will detect synthetic realities and prevent user from falling prey to complex second order effect harms. 

\section{Conclusion}
Gen-AI techniques, along with there inherent ability w.r.t translation between languages, fine-tuning between various tasks and domains \cite{desai2024opportunities} are able to overcome the challenges associated with ML/DM techniques to reduce violation w.r.t user-safety tasks.

In this survey, we provided an overview of the various user safety domains (e.g. egregious abuse, misinformation and disinformation, increase in content moderation, towards robust physical safety)  across which user safety can be violated and how Gen-AI techniques can be used to overcome those avenues. Next, we discussed how Gen-AI techniques can be used across various data modalities i.e. text, audio, video, executable binaries and images. 

We then discussed how Gen-AI techniques can be used in an adversarial setting. In particular, we discussed how these techniques can be used to pursue safety violations at scale, safety violations with human feedback, safety violations with personalized content, second order attacks and Gen-AI violations.

Lastly, we provide our opinion about the future prospects of Gen-AI techniques w.r.t user-safety. In particular, we hypothesize how Gen-AI techniques have an inherent ability w.r.t. content understanding, their ability to work as large ensembles with one model pursuing multiple tasks and lastly how Gen-AI techniques can prevent second order harm.

We believe that this work represents the first summarization of Gen-AI techniques for user-safety. In particular, we provided a deep overview of emerging technologies along with their applications to user-safety, with a focus on areas suitable for advancement. Our goal is to make sure that this work warrants immediate investment towards driving growth within the industry.

\bibliography{bibliography.bib}

\vspace{12pt}
\color{red}
\end{document}